# 5-STAR HOTEL CUSTOMER SATISFACTION ANALYSIS USING HYBRID METHODOLOGY

Yongmin Yoo[1], Yeongjoon Park[1], Dongjin Lim[1] and Deaho Seo[2]

[1]AI R&D Group, NHN, Seoul, Republic of Korea
yooyongmin91@gmail.com
yeongjoon1227@gmail.com
glow14975@gmail.com
[2]Graduate School of Information, Yonsei University
seo_daeho@naver.com

## ABSTRACT

*Due to the rapid development of non-face-to-face services due to the corona virus, commerce through the Internet, such as sales and reservations, is increasing very rapidly. Consumers also post reviews, suggestions, or judgments about goods or services on the website. The review data directly used by consumers provides positive feedback and nice impact to consumers, such as creating business value. Therefore, analysing review data is very important from a marketing point of view. Our research suggests a new way to find factors for customer satisfaction through review data. We applied a method to find factors for customer satisfaction by mixing and using the data mining technique, which is a big data analysis method, and the natural language processing technique, which is a language processing method, in our research. Unlike many studies on customer satisfaction that have been conducted in the past, our research has a novelty of the thesis by using various techniques. And as a result of the analysis, the results of our experiments were very accurate.*

## KEYWORDS

*Customer satisfaction; Data driven marketing; Text mining; Natural language processing; User-generate contents*

## 1. INTRODUCTION

Due to the amazing development of the Internet, commerce through the Internet, such as sales and reservations, is increasing very rapidly. Consumers typically post reviews, suggestions, or judgments about goods or services on the vendor's website. [1,2] These online reviews allow suppliers to obtain customer feedback. In addition, suppliers improve product or service characteristics based on feedback to provide good impact, such as creating business value [3-5]. Reviews data left by customers or surveys on satisfaction are increasingly recognized as important data for suppliers because they can reveal which aspects influence consumer loyalty and repeat purchases, increase word-of-mouth, and improve the performance of goods and services.

In particular, in the existing research on customer satisfaction, there have been many studies on guest satisfaction or dissatisfaction and positive/negative, but there are not many studies analyzing review data, which is atypical data made up of natural language. It is one of the most important tasks of big data to derive meaningful results by analyzing any data. In particular, review data with embedded meaning is difficult to utilize without detailed analysis. Therefore, more research is needed to understand consumer expectations and better leverage customer satisfaction or dissatisfaction and customer reviews.

We analyze the hotel industry, which is the most popular industry in recent years, among many fields being studied for customer satisfaction, such as food service, home appliances, and service industries. In particular, we are conducting research to analyze customer reviews of hotels, one of the industries that are booming with the end of Corona 19, and what positive/negative factors are in hotel development and customer needs.

Hotel reservation platforms using the Internet are actively used to the extent that they affect the selection of places to travel, and it is a culture for users to leave their own reviews after using a hotel, resulting in millions of reviews a day. will be registered. Since we need to analyse factors for customer satisfaction, we use data mining techniques, which are big data analysis methods. In addition, since hotel reviews are written in natural language, natural language processing techniques are also used. By using a mixture of the two techniques, we derive the positive/negative factors for the development of the hotel and the needs of the customers.

We conducted an experiment by collecting review data from several 5-star hotels called luxury hotels, and because of the experiment, we analysed the needs of customers for luxury hotels and the positive/negative factors for the development of the hotel. Compared to other products, hotels often refer to reviews when visiting, so I think our research will be of great significance.

The structure of our paper is consisting of five chapters. Chapter 1, Introduction, Chapter 2, Related Research, Chapter 3, Research Methodology, Chapter 4, Case Study, and Chapter 5, Interpretation and Conclusion.

## 2. RELATED WORK

### 2.1. Customers satisfaction

Research on customer satisfaction mainly investigates how the services (price, product quality, etc.) provided by commercial facilities affect customer satisfaction. [6,7,8] Consumer satisfaction research is being actively conducted in various fields such as restaurants, banks, and accommodation facilities. In [9,10,11] and [9], scores of food quality, service quality, and environmental conditions were surveyed on McDonald's visitors in Arab countries and how each factor affects overall satisfaction. conducted a study and found that service quality plays an important role in evaluating customer satisfaction. [10] conducted an experiment on which factors increased satisfaction with banks when using banks in Saudi Arabia. [11] investigated the effect of employee competency and service quality on customer satisfaction in the hotel industry and found that the service of front office staff had a great effect on satisfaction.

### 2.2. Review Data Textmining

Review text mining mainly conducts research using comments left by consumers after using products online, and analyses market trends and product satisfaction from comments. In addition, the analysis results are used to present the necessary elements for the product and suggest a new direction. [12,13,14] Text mining using online reviews is also being studied in various fields such as 2.1. [15,16,17] conducted a study to find and suggest new product improvement directions by analysing consumer reviews. [16] In addition, as in [15], an experiment was conducted to analyse the user's opinion to find the current trend and potential factors that can be considered in future products. [17] analysed online hotel reviews to test the effect of geographic distance and psychological distance from consumers and hotels on ratings, and short-distance travellers outperformed long-distance travellers. It has been found to give low ratings.

## 2.3. Hotel review analysis

Hotel text review analyses hotel-related information from data that exists on the Internet. Analysis is carried out in various areas, from relatively simple information such as price and equipment to complex information such as satisfaction. This information can be of great help in hotel management and improvement of service. [18,19,20] Extensive research on hotel text mining is being actively conducted [21,22,23]. [21] conducted a study on a personalized hotel recommendation system that recommends suitable hotels to users through text mining. [22] conducted a study on a system that can prevent service failures in advance by analyzing cases of hotel service failures through text mining. [23] also conducted a study on a system that recommends hotel brand positioning and competitive landscape through text mining on user-generated content.

## 2.4. Textmining

Text mining extracts necessary information from unstructured text in various ways. TFIDF [24] is a statistical numerical value indicating information about which words are important in which documents based on the frequency of occurrence of words in documents. It is used for text mining in various ways [25,26]. K-means clustering is one of the clustering methods of unsupervised learning, which groups similar data together to form a cluster. [27] Text mining has also been modified and used in various ways when analysing data characteristics [28,29]. In this case, an appropriate cluster can be formed through the Elbow method [30,31]. The Elbow method searches for the most appropriate clusters by using the section where the sum of the distances between clusters rapidly decreases as the number of clusters. LDA (Latent Dirichlet Allocation) [32,33] is one of the topic modelling techniques, and it is an algorithm that finds a topic within a document and identifies the topic of each document.

## 3. METHODOLOGY

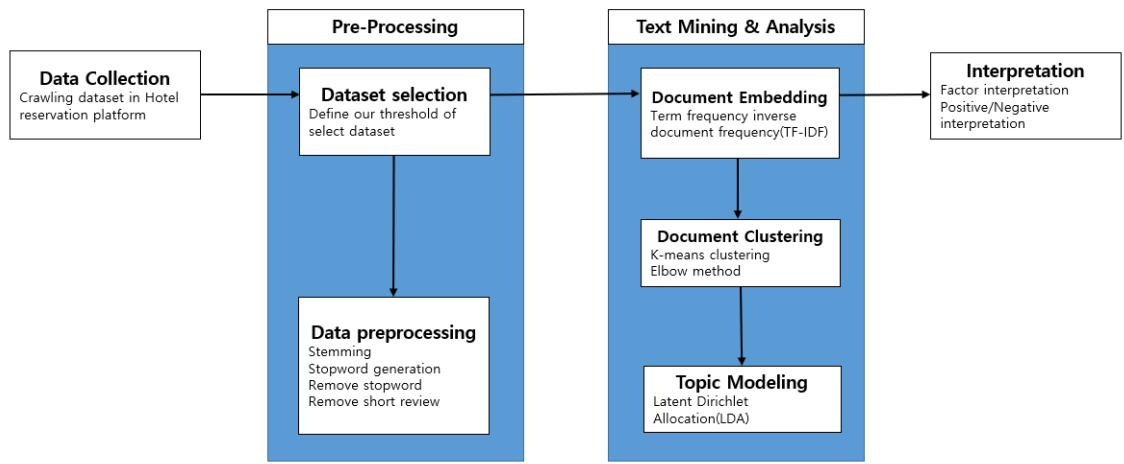

figure1. Research Flow

The various research methods used in our paper involve several steps based on previous studies. [34,35,36]. Our research method consists of four steps. The first step is crawling reviews online through the Internet, which means collecting experimental data. The second step is a data pre-processing process that transforms raw, unstructured review data into structured data for text mining. The third step is to analyse the review data using various text mining techniques. The final step is to interpret the results of text mining and text analysis.

## 3.1. Data Collection

Accommodation reservations using online platforms such as TripAdvisor, Hotels.com, Agoda, and Booking.com have become more active, allowing consumers to decide where to stay by looking at reviews written by other users before going on a trip. The review data is processed in a way that hotel users write their feelings after using the hotel along with the score. Therefore, it is a customer direct generation review on whether consumers are satisfied or dissatisfied and the factors that are important to them. In order to analyse user review data, our study collected data from five-star hotels in London provided by Booking.com. Booking.com has also been used in previous studies [34,35,36], and has the advantage of providing positive and negative reviews along with ratings as shown in Figure 2, making it convenient for pre-processing for research, and By comparison, it provides the most extensive and high-quality review data. Therefore, our research collects, analyses, and interprets the review data of booking.com.

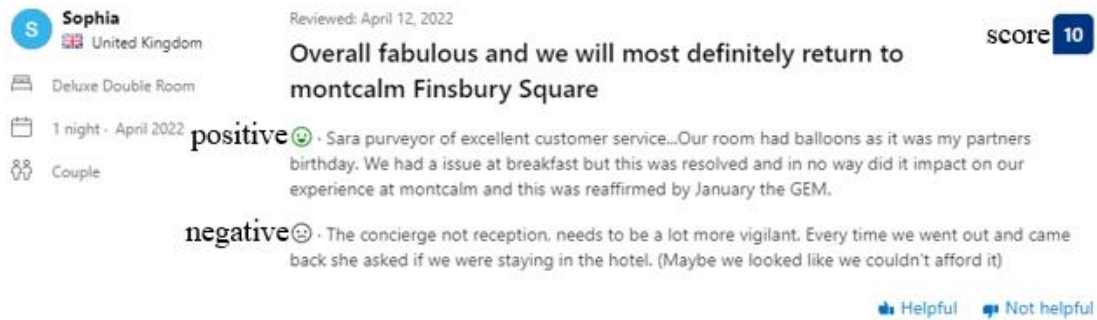

**Figure 2. Example of Review data**

## 3.2. Data pre-processing

We textmining unstructured data review data to conduct research to determine the most important factors when customers give positive/negative evaluations. Therefore, we go through several pre-processing steps to transform the unstructured data into a dataset suitable for textmining.

### 3.2.1 Dataset collection

In this paper, the entire hotel review was divided into two categories, positive and negative, and the experiment was conducted. In the case of Positive, we used review data that scored 7, 8, and 9 out of 10 out of 10 reviews. In the case of a 10-point review, which accounts for the largest portion of the overall star rating distribution, the review content was often too simple and there were many cases where there was no specific evaluation content, so it was not used. In the case of negative, data of 3 points or less were used. Review data with a score of 4 or higher often had a mixture of negative and positive content, so it was judged that it would increase the noise of the data. Finally, 9,674 positive data and 648 negative data were used.

### 3.2.2 Pre-processing

We go through four processes to delete unnecessary parts of the review dataset and conduct experiments with meta data. First, stemming all review data. The reason for stemming is that lovely and love may have the same meaning due to the characteristics of the review dataset, but if stemming is not performed, they may be treated with different meanings. In addition, there is an embedding process that maps text to vector to be applied to textmining technique in the later part, which is used to reduce the size of the index.

Second, we generate a stopword list. After calculating the term frequency of all words, we add meaningless words among the top 30 most frequent words to the stopword list. For example, add words that appear frequently but are not metadata to the stopword list, such as 'hotel', 'didn't', 'would', and 'could'.

Third, remove stopwords from the dataset based on the stopword list created in the previous step. Fourth, although expressions such as 'Very good' and 'So terrible' are expressed, reviews of less than three words often do not have a clear purpose. The reason for selecting 3 word words as the threshold is that significant contents such as 'Hotel was clean' and 'Far too loud' often start with 3 words, so the length of the data set starts with 3 words or more.

## 3.3. Textmining & Text analysis

### 3.3.1 Data Embedding
We go through the process of vectorizing the pre-processed dataset. Transform each review data into each vector using tf-idf, an embedding method traditionally used from the past for textmining.

### 3.3.2 Data Clustering
With the review data embedded in the previous step, we proceed with review data clustering. Before using the popular k-means clustering, find the optimal number of clusters using the elbow method, which finds the optimal number of clusters. Elbow method is a method to find the optimal number of clusters by looking at changes in inertia while increasing the number of clusters. It does not improve, but rather, if there are many clusters, the efficiency of the model is greatly reduced because it is meaninglessly classified into several groups, so this is a method to set the k value at the point of time when the rate of decrease is broken as the number of clusters.

**Figure2.** Algorithm *K-means clustering with elbow method*

$D \leftarrow$ Set of documents
$N \leftarrow$ Candidates of number of clustering between 1 to n
$K \leftarrow$ A number of desired cluster
**for** n=1 **to** *N* **do**
    $\mu_n \leftarrow$ some random location
    **Repeat**
    **for** d=1 **to** *D* **do**
        $Z_d \leftarrow argmin_i \|\mu_n - D_d\|$ // assign example d to closest center
    **for** n=1 **to** *N* **do**
        $\mu_n \leftarrow$ MEAN ($\{D_d : Z_d = n\}$)
    **Until** converged

### 3.3.3 Topic Modelling
We use Latent Dirichlet Allocation to extract topics representing each cluster created in the previous step. By extracting 5 topics from each cluster, and identifying the characteristics of each cluster, we find the topic words of each cluster for customer reviews. The topics extracted within each cluster will have common characteristics, and we name the cluster topic by analysing the common characteristics.

### 3.3.4 Interpretation
Our research uses natural language processing and text mining technology to analyse hotel reviews written by customers. Hotel reviews written by customers can quickly respond to customer needs and enable scientific analysis of elements necessary for hotel development and value increase.

First, we cluster the UGC for 5-star hotels by content, extract the keywords that each cluster represents through topic modelling, and check what factors are important to consider when writing customer reviews. By dividing positive and negative, it is possible to recognize the factors to increase the value of the hotel. For example, you can check the factors of good location, nice room condition and friendly staff as strengths of the hotel, and poor service, unkind staff, poor cleaning, not god facility, stay problem as weaknesses. This result means that positive/negative reviews simply do not appear as opposite results, and since it is a 5-star hotel, it can be seen from the results of the table that it is well tied to the factors included.

The conditions for being selected as a five-star hotel are very strict. Accordingly, this condition is well expressed in user reviews as well. You can check in more detail in the terms of the disadvantages rather than the advantages. For example, the terms of Poor service include terms such as breakfast, bar, and room service, but by seeing that the price is included, it can be inferred that the satisfaction is not high compared to the price (Not affordable). Also, the Not good facility and Stay problem, which do not exist in the positive, are included in the negative factor. This means that for five-star hotel users, it doesn't feel like an advantage, but it's a huge disadvantage. In other words, to maintain high user satisfaction, it is necessary to have a minimum facility, reduce noise at night, and have a good soundproofing function in the room.

First, we cluster text reviews by content and extract the keywords represented by each cluster through topic modelling to check what factors are important to consider when writing customer reviews. Through this, it is possible to recognize the factors to increase the value of the hotel. In addition, it is possible to understand what kind of trend customers have when using the hotel.

Second, we interpret the reviews of positive and negative factors separately as positive and negative. Through each interpretation of the positive and negative aspects, it is possible to establish a marketing strategy for the factors that need attention for the development of the hotel and customer satisfaction.

## 4. CASE STUDY

Our research goes through four processes. The first is to collect data, and the second is to pre-process the data. Third, we proceed with text mining and text analysis, and finally, we describe the interpretation and conclusion.

### 4.1. Data Collection

We collected hotel customer reviews from www.booking.com, a third-party booking website with the highest number of reviews and a wide range of hotel deals among many hotel booking platforms. We collected reviews of data from five-star hotels in London. London was selected because it has the most five-star hotels compared to major cities such as New York, Tokyo, and Paris. This is very demanding, and we selected it because the various conditions above affect user reviews. [33-35]

We were able to obtain user reviews from various perspectives by crawling consumer review data for a total of 164 hotels under the above conditions. As can be seen in Table 1, a total of 32,216 reviews were collected, 26,019 people visited the United Kingdom, the most nationality, and 10 points recorded the highest number of reviews with 12,775.

Data crawling used Python version 3.9.7 and Selenium version 4.0.0. The computer specs used for crawling were Intel I5-10505 without GPU and 32GB main memory.

| Nationality | Number | Percentage | Score | Number | Percentage |
|---|---|---|---|---|---|
| United Kingdom | 26,019 | 80.76% | 10 | 12,775 | 39.65% |
| United States of America | 1002 | 3.11% | 9 | 7619 | 23.65% |
| Australia | 452 | 1.40% | 8 | 4672 | 14.50% |
| Ireland | 395 | 1.23% | 3 | 471 | 1.46% |
| United Arab Emirates | 344 | 1.07% | 2 | 232 | 0.72% |
| | | | 1 | 354 | 1.10% |
| | | | **Total Sample** | 32,216 | |

**Table1. Dataset configuration**

## 4.2. Data Pre-processing

In order to mine and analyze the data collected in the previous step, we go through two big steps of pre-processing the data. First, we select the data to be used for the experiment. Second, pre-processing is performed to leave only the meta data necessary for analysis. Pre-processing is performed in four steps: stemming, generating a list of stopwords, deleting stopwords, and deleting data less than 3 words.

### 4.2.1 Data selection

The collected data has positive and negative reviews in one review and a rating (star rating). Among them, we select the positive and negative reviews to be used in the study in the way we defined. For Positive, we used review data that scored 7, 8, or 9 out of 10 out of 10 reviews. In the case of a 10-point review, which accounts for the largest portion of the overall star rating distribution, we did not use it because the review content was often too simple and there were many cases where there was no specific evaluation content. For Negative, we used data with a score of 3 or less. Review data with a score of 4 or higher often had a mixture of negative and positive content, so it was judged that it would increase the noise of the data. Finally, 9,674 positive data and 648 negative data were used.

### 4.2.2 Pre-processing

We delete unnecessary parts of the review dataset and go through 4 processes as shown in Fig3. to proceed with the experiment with a significant dataset.

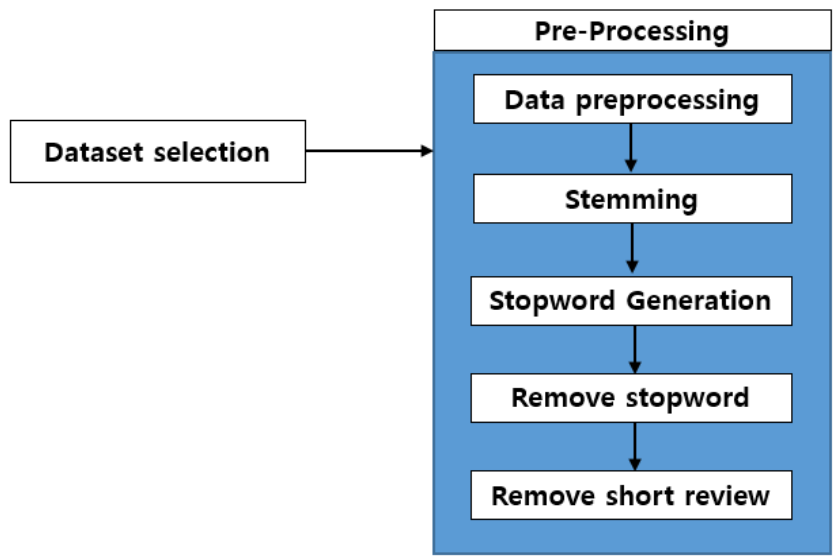

**Figure 3. Pre-processing Flow**

First, we tokenize all datasets by word unit and then proceed with stemming. If you proceed with Stemming, the review data will change to the form shown in Fig4.

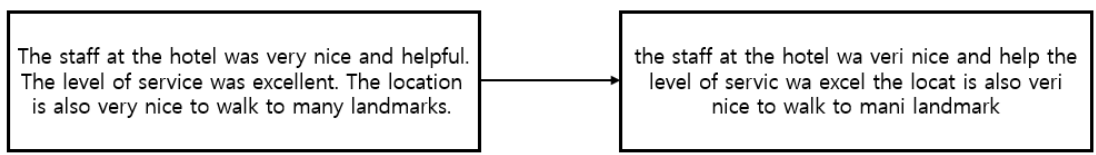

**Figure 4. Example of Stemming**

Second, we create a stopword list. By default, stopword in the NLTK library is used. However, due to the nature of hotel reviews, we propose a new method to create a stopword list due to unnecessary repeated words. After calculating the term frequency of all words, we add meaningless words among the top 30 most frequent words to the stopword list.

Third, remove the stopword based on the stopword list created earlier. After these three steps, the review data is processed in the form shown in Fig5.

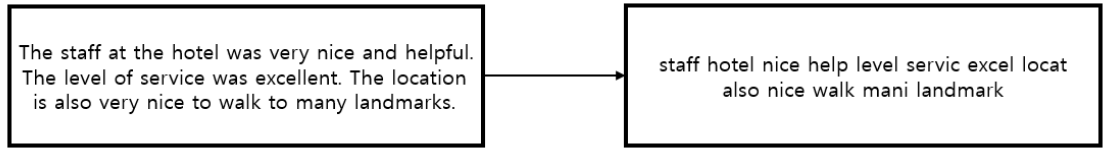

**Figure 5. Example of remove stopword**

Finally, we delete reviews with less than 3 words. The table below shows the number of pre-processed datasets and positive/negative numbers.

**Table 2. Number of data sets after pre-processing**

|  | # of Before Pre-processing | # of After Pre-processing |
|---|---|---|
| **Positive** | 10,586 | 9,674 |
| **Negative** | 863 | 648 |

## *4.3 Textmining & Text analysis*

We proceed with textmining for text analysis with the dataset that has been preprocessed in the previous step. The textmining process goes through the following three steps.

First, we go through the process of vectorizing a dataset of characters. Transform each review data into each vector using tf-idf, an embedding method traditionally used from the past for textmining.

Second, we cluster reviews of the same content. The clustering method uses k-means. Since we do not know the appropriate number of clusters prior to use, we first find the optimal number of clusters using the Elbow method as shown in the figure below. In the case of positive review, it was found that it was optimal to group it into three topics and in the case of negative review into five topics.

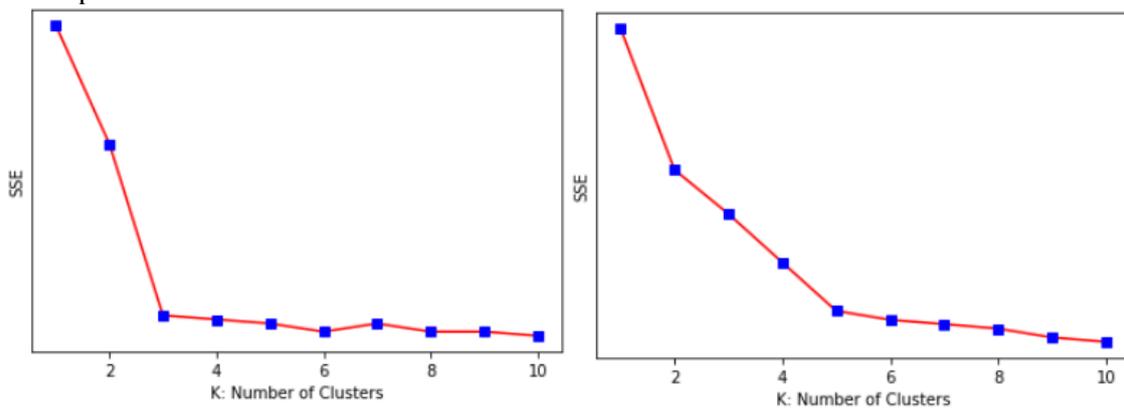

**Figure 6. Number of clusters**

After that, each review is clustered using k-means clustering according to the number of clusters previously determined. In the case of positive review, three topics were grouped, and in the case of negative review, five topics were grouped.

Third, we extract five topics from each cluster through LDA to find out what topics and elements each cluster of positive/negative data are clustered with. However, we removed adjectives such as 'amaze', 'terrific', and 'awesome' when they were extracted as terms. The reason is that since the analysis started with data divided into positive and negative in the past, it is treated as a word that is not necessary to know the customer satisfaction factor or the development direction of the hotel. The table below shows the extraction results for each cluster of positive/negative data.

Table 3. Factors regarding customer positive reviews

| Cluster | Topic | Term |
|---|---|---|
| C1 | Good location | 'locat','excel locat', 'servic', 'locat excel', 'staff excel' |
| C2 | Nice room condition | 'room', 'clean', 'view', 'comfort', 'neat' |
| C3 | Friendly staff | 'staff', 'friendli', 'staff friendli', 'help', 'friendli staff'' |

Table 4. Factors regarding customer negative reviews

| Cluster | Topic | Term |
|---|---|---|
| C1 | Poor service | 'servic', 'breakfast', 'bar', 'room servic', 'price' |
| C2 | Unkind staff | 'staff', 'check', 'told', 'rude', 'manag' |
| C3 | Poor cleaning | 'dirti', 'bed', 'bathroom', 'clean', 'floor' |
| C4 | Not good facility | 'book', 'spa', 'pool', 'us', 'facil' |
| C5 | Stay problem | 'night', 'sleep', 'noi', 'stay', 'bed', |

## 5. INTERPRETATION & CONCLUSION

Our research uses natural language processing and text mining technology to analyse hotel reviews written by customers. Hotel reviews written by customers can respond quickly to customer needs and complaints and enable scientific analysis of factors necessary for hotel development and value increase.

First, we cluster review data, which is a type of CGC (Customer Generated Content), for a 5-star hotel by content, and extract the keywords indicated by each cluster through topic modeling, a factor that is important to consider when writing customer reviews. check which part is there. By dividing positive and negative, it is possible to recognize the factors to increase the value of the hotel. For example, Table. You can check the factors of good location, nice room condition and friendly staff as the strengths of the hotel, and poor service, unkind staff, poor cleaning, not god facility, and stay problem as weaknesses. This result means that when customers evaluate a hotel, there is a difference between the reasons for determining the positive and the negative. Also, since it is a five-star hotel, it can be seen from the results of the table that it is well tied to the factors included.

The conditions for being selected as a five-star hotel are very strict. Accordingly, this condition is well expressed in user reviews as well. In particular, you can check in more detail in the terms of the disadvantages rather than the advantages. For example, the terms of Poor service include terms such as breakfast, bar, and room service, but by seeing that the price is included, it can be inferred that the satisfaction is not high compared to the price are (Not affordable). Also, the Not good facility and Stay problem, which do not exist in the positive, are included in the negative factor. For 5-star hotel users, it means that even if factors such as facility are satisfactory, they do not feel it as a big advantage, but it comes as a big disadvantage. In other words, to maintain high user satisfaction, it is necessary to have a minimum facility, reduce noise at night, and have a good soundproofing function in the room.

As a result of the experiment, we found customer needs and complaints from customer generate contents by using a mixture of data mining methods and natural language processing methods. There was. Our method is automatic based on scientific method, so anyone can use it comfortably. In addition, because it is not a black box algorithm that is difficult to explain like the Deep Learning Algorithm, the results are easy to explain, which has the advantage of logically persuading the hotel manager and marketing team to establish future strategies.

In this study, we collected CGCs from five-star hotels in downtown London, analyzed them using NLP and data mining techniques, and interpreted the results. These interpretations are good points for solving problems and finding direction from the hotel point of view. From the point of view of customers who select and use one hotel, if the analysis is conducted on individual hotels rather than collecting CGC in one place, the hotel is selected by factors (such as review, score, etc.) that are important to customers when selecting a hotel. It is expected that it can be used as one of the methods of hotel selection such as analysis. In the future, it is expected that our technique can be applied to various data such as product reviews as well as hotel reviews.